\documentclass[a4paper, 10pt, conference]{ieeeconf}      

\IEEEoverridecommandlockouts                              

\usepackage{graphics}
\usepackage{amsmath,amssymb}
\usepackage[table]{xcolor}            
\usepackage{subcaption}               
\usepackage{siunitx}                  
\usepackage[textsize=tiny]{todonotes} 
\usepackage{lipsum}                   
\usepackage{dblfloatfix}              
\usepackage{tabu}                     

\usepackage{url}

\makeatletter
\let\NAT@parse\undefined
\makeatother
\usepackage{hyperref}

\newcommand{\fig}[1]{Fig.~\ref{#1}}

\newcommand{\eq}[1]{(\ref{#1})}
\newcommand{\tab}[1]{Table~\ref{#1}}

\newcommand{\sect}[1]{Section~\ref{#1}}

\newcommand{\vect}[1]{\boldsymbol{#1}}
\newcommand{\mat}[1]{\mathbf{#1}}

\newcommand\copyrighttext{%
    \footnotesize \copyright{ }2020 IEEE. Personal use of this material is permitted. Permission from IEEE must be obtained for all other uses, in any current or future media, including reprinting/republishing this material for advertising or promotional purposes, creating new collective works, for resale or redistribution to servers or lists, or reuse of any copyrighted component of this work in other works.}
\newcommand\copyrightnotice{%
    \begin{tikzpicture}[remember picture,overlay]
    \node[anchor=south,yshift=10pt,xshift=7pt] at (current page.south) {\parbox{\dimexpr\textwidth-\fboxsep-\fboxrule\relax}{\copyrighttext}};
    \end{tikzpicture}%
}

\title{\LARGE \bf
  A Sim2Real Deep Learning Approach for the\\
  Transformation of Images from Multiple Vehicle-Mounted Cameras\\
  to a Semantically Segmented Image in Bird's Eye View*
}

\author{Lennart Reiher$^{1}$ and Bastian Lampe$^{1}$, Lutz Eckstein$^{2}$
  \thanks{*This research is accomplished within the project "UNICAR\textit{agil}"~(FKZ~16EMO0289). We acknowledge the financial support for the project by the Federal Ministry of	Education and Research of Germany (BMBF).}
  \thanks{$^{1}$The authors contributed equally to this work. They are with the Institute for Automotive Engineering~(ika), 
  RWTH Aachen	University, 52074 Aachen, Germany.
  {\tt\small \{firstname.lastname\}@ika.rwth-aachen.de}}
  \thanks{$^{2}$Lutz Eckstein is head of the Institute for Automotive Engineering~(ika),
  RWTH Aachen	University, 52074 Aachen, Germany.
	{\tt\small lutz.eckstein@ika.rwth-aachen.de}}
}

\begin{document}

\maketitle
\thispagestyle{empty}
\pagestyle{empty}
\copyrightnotice

\begin{abstract}
  Accurate environment perception is essential for automated driving. When using monocular cameras, the distance estimation of elements in the environment poses a major challenge. Distances can be more easily estimated when the camera perspective is transformed to a bird's eye view (BEV). For flat surfaces, \textit{Inverse Perspective Mapping} (IPM) can accurately transform images to a BEV. Three-dimensional objects such as vehicles and vulnerable road users are distorted by this transformation making it difficult to estimate their position relative to the sensor. This paper describes a methodology to obtain a corrected 360\(^{\circ}\) BEV image given images from multiple vehicle-mounted cameras. The corrected BEV image is segmented into semantic classes and includes a prediction of occluded areas. The neural network approach does not rely on manually labeled data, but is trained on a synthetic dataset in such a way that it generalizes well to real-world data. By using semantically segmented images as input, we reduce the reality gap between simulated and real-world data and are able to show that our method can be successfully applied in the real world. Extensive experiments conducted on the synthetic data demonstrate the superiority of our approach compared to IPM. Source code and datasets are available at \url{https://github.com/ika-rwth-aachen/Cam2BEV}.
\end{abstract}

\section{Introduction}

In recent years, the development of automated vehicles~(AVs) has received substantial attention from both research and industry. One of the key elements of automated driving is the accurate perception of an AV's environment. It is essential for planning safe and efficient behavior.

Different types of environment representations can be computed, e.g.\ object lists or occupancy grids. Both require information on the world coordinates of elements in the environment. Among the different types of sensors commonly used to achieve an understanding of the environment, cameras are popular due to low cost and well-established computer vision techniques. Since monocular cameras can only provide information on locations in the image plane, a perspective transformation can be applied to images that results in a top-down or bird's eye view~(BEV). It is an approximation of the same scene as seen from a perspective in which the image plane aligns with the ground plane in front of the camera. The method used for transforming camera images to BEV is commonly referred to as \textit{Inverse Perspective Mapping}~(IPM)~\cite{MallotEtAl_InversePerspectiveMapping_1991}.

IPM assumes the world to be flat. Any three-dimensional object and changing road elevations violate this assumption. Mapping all pixels to a flat plane thus results in strong visual distortions of such objects. This impedes our goal of accurately locating objects such as other vehicles and vulnerable road users in the vehicle's environment. For this reason, images transformed through IPM often only serve as input to algorithms for lane detection or free space computation, for which the flat world assumption is often reasonable~\cite{BarHillelEtAl_RecentProgressRoad_2014}.

Even if errors introduced by IPM could be corrected, we are left with the task of detecting objects in the BEV. Deep learning approaches have proven to be powerful for tasks like semantic segmentation of images but usually require vast amounts of manually labeled data. Simulations can provide BEV images and their corresponding labels but suffer from the so-called reality gap: BEV images computed by a virtual camera in a simulated environment are rather dissimilar to e.g.\ a drone image captured above a vehicle in the real world, mostly due to unrealistic textures in the simulation. The generalization from a complex task learned in a simulation to the real world has therefore proven to be difficult so far. In order to reduce the reality gap, many approaches thus aim at making simulated data more realistic, e.g.~\cite{ZhaoEtAl_MultiSourceDomain_2019}. 

\begin{figure}[t]
  \captionsetup[subfigure]{skip=0pt}
  \begin{subfigure}[t]{0.24\linewidth}
    \captionsetup{belowskip=6pt}
    \caption*{Front camera}
    \includegraphics[width=\textwidth, height=0.5\textwidth]{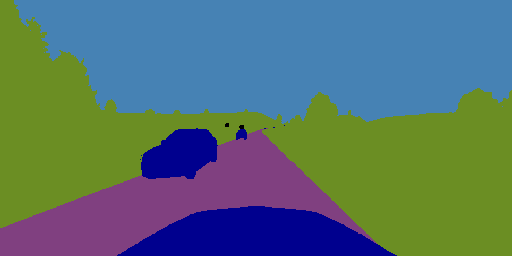}
  \end{subfigure}
  \hfill
  \begin{subfigure}[t]{0.24\linewidth}
    \captionsetup{belowskip=6pt}
    \caption*{Rear camera}
    \includegraphics[width=\textwidth, height=0.5\textwidth]{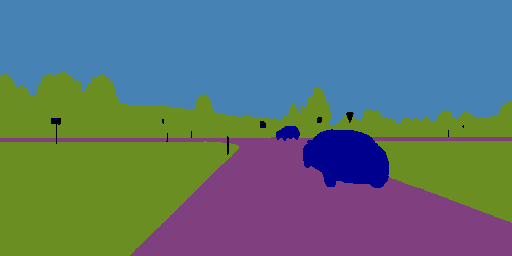}
  \end{subfigure}
  \hfill
  \begin{subfigure}[t]{0.24\linewidth}
    \captionsetup{belowskip=6pt}
    \caption*{Left camera}
    \includegraphics[width=\textwidth, height=0.5\textwidth]{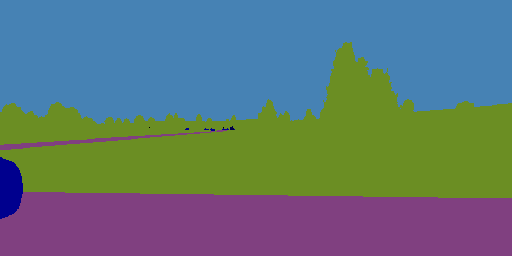}
  \end{subfigure}
  \hfill
  \begin{subfigure}[t]{0.24\linewidth}
    \captionsetup{belowskip=6pt}
    \caption*{Right camera}
    \includegraphics[width=\textwidth, height=0.5\textwidth]{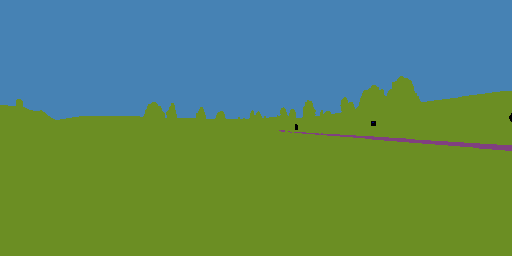}
  \end{subfigure}%

  \vspace{0.25\baselineskip}
  \captionsetup[subfigure]{skip=2pt}
  \begin{subfigure}[b]{0.4925\linewidth}
    \includegraphics[width=\textwidth, height=0.5\textwidth]{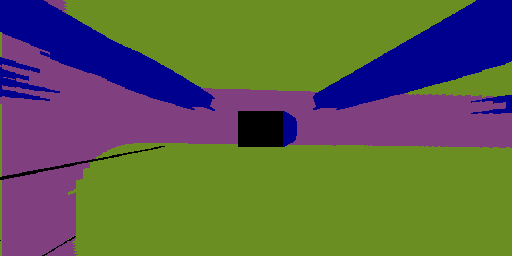}
    \caption*{Homography}
  \end{subfigure}
  \hfill
  \begin{subfigure}[b]{0.4925\linewidth}
    \includegraphics[width=\textwidth, height=0.5\textwidth]{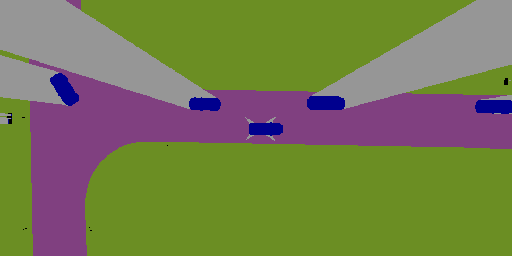}
    \caption*{Ground Truth BEV}
  \end{subfigure}%
  \caption{A homography can be applied to the four semantically segmented images from vehicle-mounted cameras to transform them to BEV. Our approach involves learning to compute an accurate BEV image without visual distortions.}
  \label{fig:TeaserImage}
\end{figure}

In this paper, we propose a methodology to obtain BEV images that are not subject to the errors introduced by the flatness-assumption underlying IPM. Instead of trying to make simulated images look more realistic, we remove mostly unnecessary texture from real-world data by computing semantically segmented camera images. We show how their use as input to our algorithm allows us to train a neural network on synthetic data only, while still being able to successfully perform the desired task on real-world data. With semantically segmented input, the algorithm has access to class information and is thus able to incorporate these into the correction of images produced by IPM. The output is a semantically segmented BEV of the input scene. Since the object shapes are preserved, the output can not only be used for determining free space but also to locate dynamic objects. In addition, the semantically segmented BEV images contain a color-coding for unknown areas, which are occluded in the original camera image. The image obtained through IPM and the desired ground truth BEV image are displayed in~\fig{fig:TeaserImage}.

The main contributions of this work are as follows:
\begin{itemize}
  \item We propose a methodology capable of transforming the images of multiple vehicle-mounted cameras to semantically segmented images in BEV.
  \item We design and compare two variations of our methodology using different neural network architectures, one of which we specifically design for the task.
  \item We design the process in such a way that no manual labeling of BEV images is required for training our neural network-based models.
  \item We show a successful real-world application of the trained models.
\end{itemize}

\section{Related Work}

Numerous works of literature address the perspective transformation to BEV. In the automotive context, both~\cite{SungEtAl_DevelopmentImageSynthesis_2012} and~\cite{ZhangEtAl_SurroundViewCamera_2014} deal with the synthesized transformation of multiple camera images to a top-down surround view. Most works are geometry-based and focus on an accurate depiction of the ground level.

Only few works combine the transformation to BEV with the task of scene understanding. However, object detection can give clues on an object's geometry, from which the transformation could benefit. Recently, the deep learning approaches presented below have shown how complex neural networks can aid in improving the classical IPM technique and contribute to environment perception.

The focus of~\cite{BrulsEtAl_RightAngledPerspective_2019} and~\cite{ZhuEtAl_GenerativeAdversarialFrontal_2019} is to correct the errors introduced by the IPM approach. Dynamic and three-dimensional objects are sought to be removed in the transformed BEV achieved by~\cite{BrulsEtAl_RightAngledPerspective_2019} to improve road scene understanding. In contrast, the method proposed in~\cite{ZhuEtAl_GenerativeAdversarialFrontal_2019} aims to synthesize an accurate BEV representation of an entire road scene as seen through a front-facing camera, including dynamic objects. Due to the generative nature of the underlying task, both methods employ Generative Adversarial Networks~\cite{Schmidhuber_MakingWorldDifferentiable_1990, GoodfellowEtAl_GenerativeAdversarialNets_2014}.

Palazzi~et~al.~\cite{PalazziEtAl_LearningMapVehicles_2017} present the prediction of vehicle bounding boxes in BEV from the images of a front-facing camera. Roddick~et~al.~\cite{RoddickEtAl_OrthographicFeatureTransform_2019} demonstrate advanced object detection in computing three-dimensional bounding boxes by using an in-network orthographic feature transform to a three-dimensional discretization of space.

A semantic road understanding in a top-down frame leading to a coarse and static semantic map is achieved in~\cite{SenguptaEtAl_AutomaticDenseVisual_2012}. Similar to~\cite{BrulsEtAl_RightAngledPerspective_2019}, this approach tries to remove dynamic traffic participants.

To the best of our knowledge, the only source pursuing the idea of directly transforming multiple semantically segmented images to BEV is a blog article~\cite{Dziubinski_SemanticSegmentationSemantic_2019}. It lacks detailed testing and an application to real-world data though. The designed neural network is a fully-convolutional autoencoder and has multiple weaknesses, e.g.\ the range of an accurate object detection is relatively low.

\section{Methodology}\label{sec:Methodology}

We base our methodology on the use of a Convolutional Neural Network~(CNN), a class of deep neural networks commonly used for image analysis. Most popular CNNs process only one input image. In order to fuse images from multiple cameras mounted on a vehicle, a single-input network could take as input multiple images concatenated along their channel dimension. However, for the task at hand, this would result in spatial inconsistency between input and output images. Convolutional layers operate locally, i.e.\ information in particular parts of the input are mapped to approximately the same part of the output. An end-to-end learning approach for the presented problem however needs to be able to handle images from multiple viewpoints. This suggests the need for an additional mechanism.

IPM certainly introduces errors, but the technique is capable of producing an image at least similar to a ground truth BEV image. Due to this similarity, it seems reasonable to incorporate IPM as a mechanism to provide better spatial consistency between input and output images. The image resulting from IPM is also used as an intermediate guiding view in~\cite{BrulsEtAl_RightAngledPerspective_2019} and~\cite{ZhuEtAl_GenerativeAdversarialFrontal_2019}. In the following, we present two variations of our neural network-based methodology that both include the application of IPM. Before introducing the two neural network architectures, the applied data preprocessing techniques are explained in detail.

\subsection{Dealing with Occlusions}

When only considering the input domain and the desired output for this task, one difficulty immediately becomes apparent: traffic participants and static obstacles may occlude parts of the environment making predictions for those areas in a BEV image mostly impossible. As an example, such occlusions would occur when driving behind a truck: what is happening in front of the truck cannot reliably be determined only from vehicle-mounted camera images.

\begin{figure}[t]
  \centering
  \includegraphics[width=\linewidth]{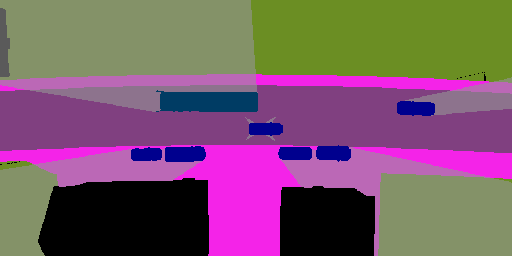}
  \caption{The original ground truth image is overlayed by the modified label including the \textit{occlusion} class (gray shade). The \textit{cars} (in blue) and the \textit{bus} (in dark turquoise) occlude the ground behind them. The \textit{buildings} behind the parked passenger cars are still visible, while the \textit{bus} also blocks sight on the \textit{building} in the top left corner of the view. Although the parked vehicles partially occlude each other, they remain completely visible.}
  \label{fig:occlusion}
\end{figure}

In order to formulate a well-posed problem, an additional semantic class needs to be introduced for areas in BEV, which are occluded in the camera perspectives. This class is introduced to the ground truth label images in a preprocessing step. For each vehicle camera, virtual rays are cast from its mount position to the edges of the semantically segmented ground truth BEV image. The rays are only cast to edge pixels that lie within the specific camera's field of view. All pixels along these rays are processed to determine their occlusion state according to the following rules:
\begin{itemize}
  \item some semantic classes always block sight \\(e.g.\ \textit{building}, \textit{truck});
  \item some semantic classes never block sight (e.g.\ \textit{road});
  \item \textit{cars} block sight, except on taller objects behind them (e.g.\ \textit{truck}, \textit{bus});
  \item partially occluded objects remain completely visible;
  \item objects are only labeled as \textit{occluded} if they are occluded in all camera perspectives.
\end{itemize}
A ground truth BEV image modified according to these rules is showcased in~\fig{fig:occlusion}.

\subsection{Projective Preprocessing}\label{sec:ProjectivePreprocessing}

As part of the incorporation of the IPM technique into our methods, the homographies, i.e.\ the projective transformations between vehicle camera frames and BEV are derived. The determination of the correct homography matrix involves intrinsic and extrinsic camera parameters and shall be briefly described below.

The relationship between homogeneous world coordinates \( \vect{x}_w \in \mathbb{R}^4 \) and homogeneous image coordinates \( \vect{x}_i \in \mathbb{R}^3 \) is given by the projection matrix \( \mat{P} \in \mathbb{R}^{3 \times 4} \) as
\begin{equation}
  \vect{x}_i = \mat{P} \vect{x}_w \,.
  \label{eq:World2ImageProjection}
\end{equation}
The projection matrix encodes the camera's intrinsic parameters (e.g., focal length) in a matrix \( \mat{K} \) and extrinsics (rotation \( \mat{R} \) and translation \( \vect{t} \) w.r.t.\ the world frame):
\begin{equation}
  \mat{P} = \mat{K} \left[ \mat{R} \vert \vect{t} \right] \,.
  \label{eq:ProjectionMatrix}
\end{equation}
Assuming there exists a transformation \( \mat{M} \in \mathbb{R}^{4 \times 3} \) from the road plane \( \vect{x}_r \in \mathbb{R}^3 \) to the world frame, s.t.\
\begin{equation}
  \vect{x}_w = \mat{M} \vect{x}_r \,,
  \label{eq:Road2WorldProjection}
\end{equation}
we obtain a transformation from image coordinates to the road plane:
\begin{equation}
  \vect{x}_r = \left( \mat{P} \mat{M} \right)^{-1} \vect{x}_i \,.
\end{equation}
Note that \eq{eq:World2ImageProjection} is generally not invertible, as infinitely many world points correspond to the same image pixel. The assumption of a planar surface, encoded in \( \mat{M} \), makes it possible to construct the invertible matrix \( \left( \mat{P} \mat{M} \right) \).

In order to determine \( \mat{P} \) for real-world cameras, camera calibration methods~\cite{KaehlerBradski_LearningOpenCVComputer_2017} can be used.

As a preprocessing step to the first variation of our approach~(\sect{sec:SingleInputModel}), IPM is applied to all images from the vehicle cameras. The transformation is set up to capture the same field of view as the ground truth BEV image. As this area is only covered by the union of all camera images, they are first separately transformed via IPM and then merged into a single image, hereafter called the \textit{homography image}. Pixels in overlapping areas, i.e.\ areas visible from two cameras, are chosen arbitrarily from one of the transformed images.

\subsection{Variation 1: Single-Input Model}\label{sec:SingleInputModel}

As the first variation of our approach, we propose to pre-compute the homography image as presented in~\sect{sec:ProjectivePreprocessing} in order to bridge a large part of the gap between camera views and BEV. Hereby we provide, to some extent, spatial consistency between neural network input and output. The network's task then is to correct the errors introduced by IPM.

To the best of our knowledge, there exist no single-input neural network architectures, which specifically target the problem at hand. However, since the homography image and the desired target output image cover the same spatial region, we propose to use existing CNNs for image processing, which have proven successful at other tasks such as semantic segmentation.

We choose \textit{DeepLabv3+} as the architecture for our proposed single-network-input method. \textit{DeepLabv3+} as presented in~\cite{ChenEtAl_EncoderDecoderAtrousSeparable_2018} is a state-of-the-art CNN for semantic image segmentation. With \textit{MobileNetV2}~\cite{SandlerEtAl_MobileNetV2InvertedResiduals_2018} and \textit{Xception}~\cite{Chollet_XceptionDeepLearning_2017}, two different network backbones are tested. The resulting neural networks have approximately 2.1M and 41M trainable parameters.

\subsection{Variation 2: Multi-Input Model}\label{sec:MultiInputModel}

In contrast to the first network architecture presented in~\sect{sec:SingleInputModel}, we propose a second neural network that processes all non-transformed images from the vehicle cameras as input. It therefore extracts features in the non-transformed camera views and is thus not fully subject to the errors introduced by the IPM. As a way to deal with the problem of spatial inconsistency, we integrate projective transformations into the network.

In order to build an architecture for multiple input and one output image, we propose to extend an existing CNN to multiple input streams with a fusion of said streams inside. Due to its simplicity and thus easy extensibility, we choose the popular semantic segmentation architecture \textit{U-Net}~\cite{RonnebergerEtAl_UNetConvolutionalNetworks_2015} as the basis for the extensions presented in the following.

\begin{figure*}[t]
  \centering
  \includegraphics[width=0.65\textwidth]{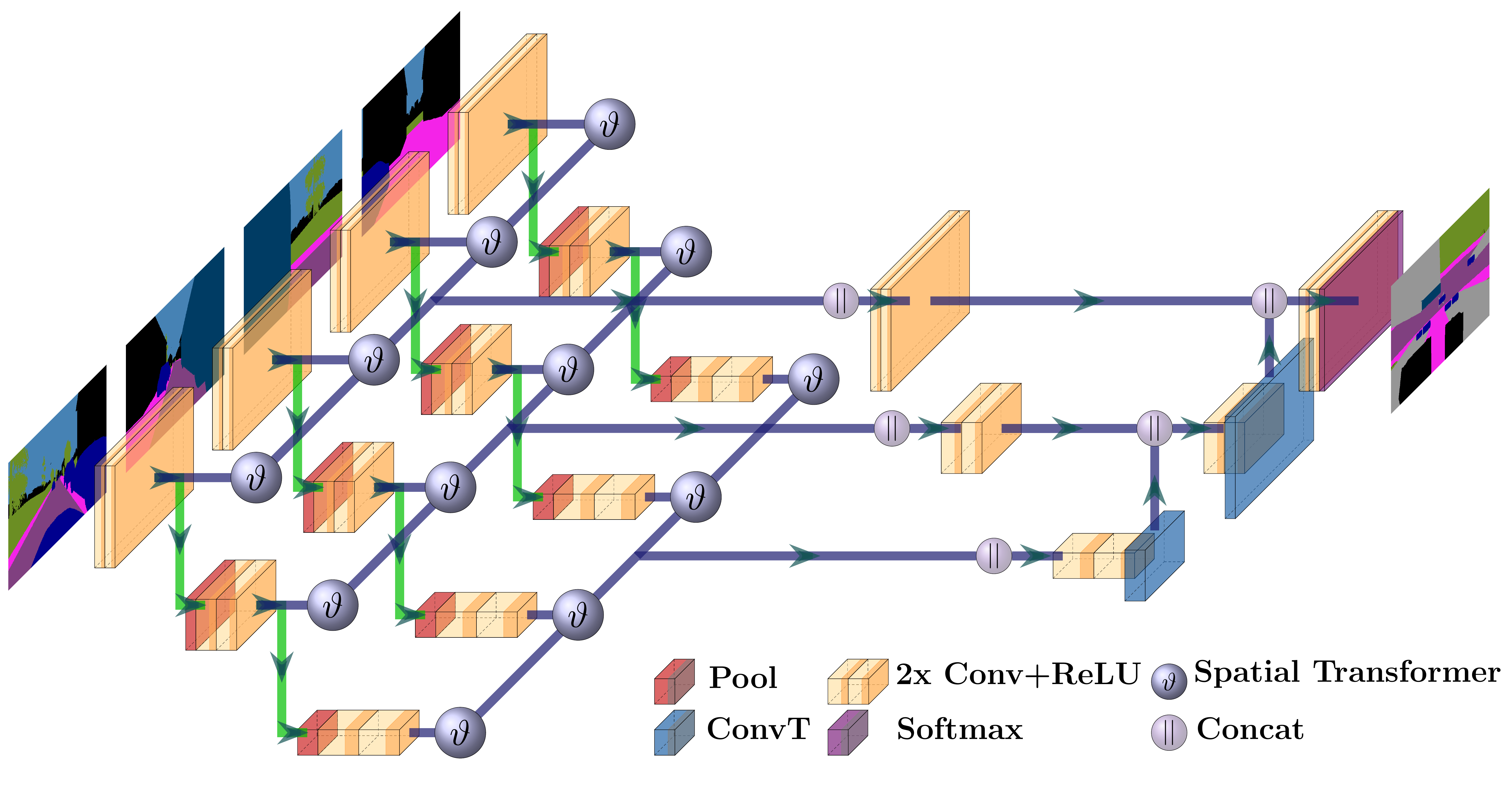}
  \captionsetup{width=0.65\textwidth}
  \caption{The \textit{uNetXST} architecture has separate encoder paths for each input image (green paths). As part of the skip-connection on each scale level (violet paths), feature maps are projectively transformed (\( \vartheta \)-block), concatenated with the other input streams (\( \vert\vert \)-block), convoluted, and finally concatenated with upsampled output of the decoder path. This illustration shows a network with only two pooling and two upsampling layers, the actual trained network contains four, respectively.}
  \label{fig:uNetX}
\end{figure*}

\begin{figure}[!b]
  \centering
  \includegraphics[width=0.75\linewidth]{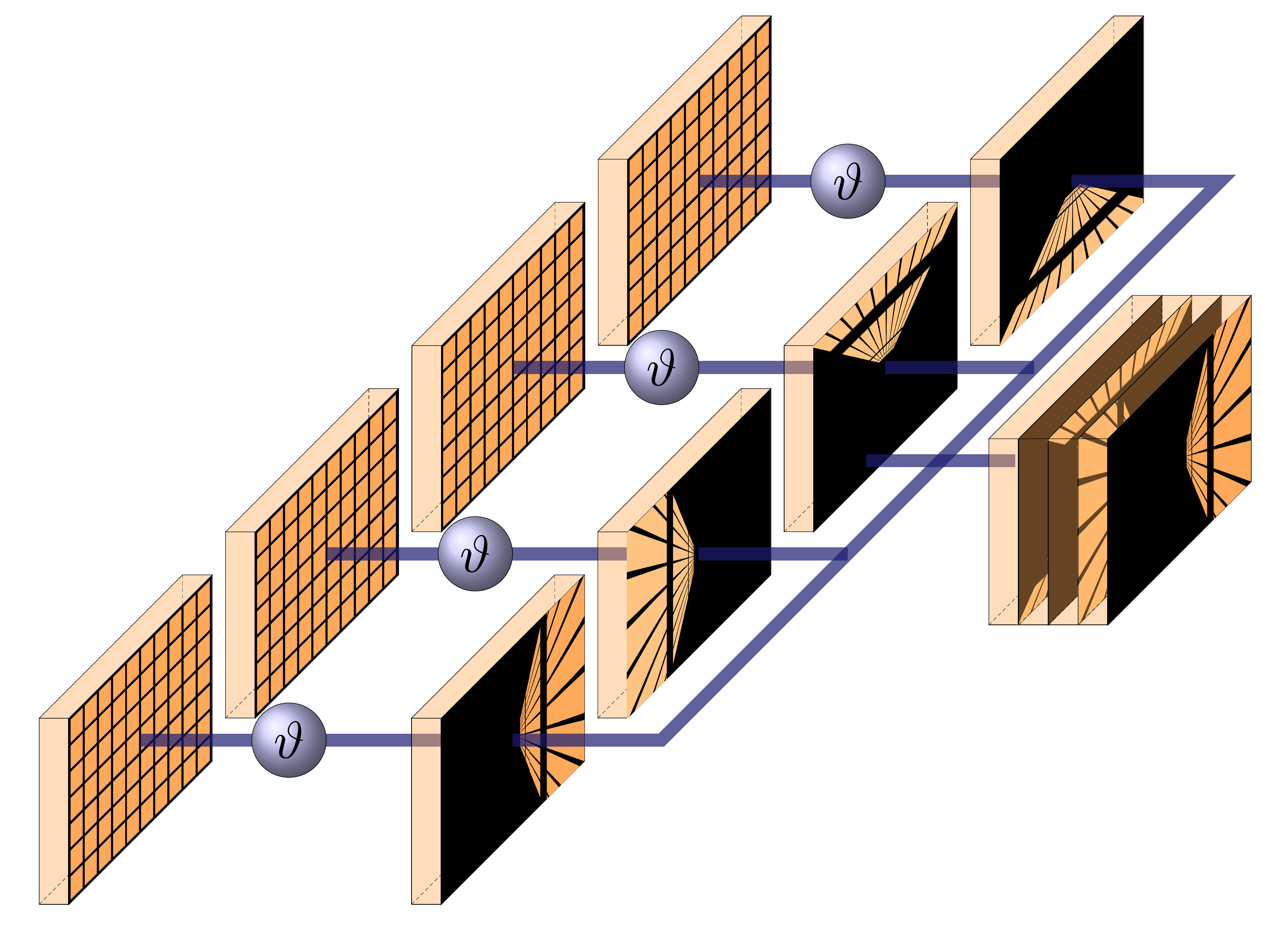}
  \caption{The \( \vartheta \)-block resembles a \textit{Spatial Transformer} unit. Input feature maps from preceding convolutional layers (orange grid layers) are projectively transformed by the homographies obtained through IPM. The transformation differs between the input streams for the different cameras. Spatial consistency is established, since the transformed feature maps all capture the same field of view as the ground truth BEV. The transformed feature maps are then concatenated into a single feature map (cf.\ \( \vert\vert \)-block).}
  \label{fig:SpatialTransformer}
\end{figure}

The base architecture consists of a convolutional encoder and decoder path based on successive pooling and upsampling, respectively. Additionally, high-resolution features from the encoder side are combined with upsampled outputs on the decoder side via skip-connections on each scale. \fig{fig:uNetX} shows the architecture including the two extensions that are introduced in order to handle multiple input images and add spatial consistency:
\begin{enumerate}
  \item The encoder path is separately replicated for each input image. For every scale, features from each input stream are concatenated and convoluted to build the skip-connection to the single decoder path.
  \item Before concatenating the input streams, \textit{Spatial Transformer}~\cite{JaderbergEtAl_SpatialTransformerNetworks_2015} units projectively transform the feature maps using the fixed homography as obtained by IPM. These transformers are explained in more detail in~\fig{fig:SpatialTransformer}.
\end{enumerate}

The neural network is named \textit{uNetXST} due to its extension to arbitrarily many inputs and the \textit{Spatial Transformer} units. It contains approximately 9.6M trainable parameters.

\setcounter{table}{1}
\begin{table*}[!b]
  \caption{Class IoU Scores (\%) on the Validation Set}
  \setlength{\tabcolsep}{2pt}
  \begin{center}
    \begin{tabu} to \textwidth{l X[c] X[c] X[c] X[c] X[c] X[c] X[c] X[c] X[c] X[c]}
      \textbf{Model} & \cellcolor[RGB]{128,64,128}\textcolor{white}{\textbf{Road}} & \cellcolor[RGB]{244,35,232}\textcolor{white}{\textbf{Sidewalk}} & \cellcolor[RGB]{220,20,60}\textcolor{white}{\textbf{Person}} & \cellcolor[RGB]{0,0,142}\textcolor{white}{\textbf{Car}} & \cellcolor[RGB]{0,0,70}\textcolor{white}{\textbf{Truck}} & \cellcolor[RGB]{0,60,100}\textcolor{white}{\textbf{Bus}} & \cellcolor[RGB]{255,0,0}\textcolor{white}{\textbf{Bike}} & \cellcolor[RGB]{0,0,0}\textcolor{white}{\textbf{Obstacle}} & \cellcolor[RGB]{107,142,35}\textcolor{white}{\textbf{Vegetation}} & \cellcolor[RGB]{150,150,150}\textcolor{white}{\textbf{Occluded}}\\
      \hline
      \\[-1em]
      uNetXST & \textbf{\num{98.10}} & \num{93.36} & \textbf{\num{13.56}} & \textbf{\num{80.90}} & \num{65.82} & \num{62.10} & \textbf{\num{32.43}} & \num{88.99} & \textbf{\num{97.27}} & \textbf{\num{86.62}}\\
      DL Xception & \num{98.06} & \textbf{\num{94.02}} & \num{6.93} & \num{80.21} & \textbf{\num{65.94}} & \textbf{\num{65.98}} & \num{30.80} & \textbf{\num{89.05}} & \num{97.09} & \num{85.42}\\
      DL MobileNetV2 & \num{96.93} & \num{91.51} & \num{0.00} & \num{76.05} & \num{60.33} & \num{64.92} & \num{16.79} & \num{85.83} & \num{96.28} & \num{77.31}\\
      DL Xception* & \num{96.60} & \num{88.81} & \num{0.20} & \num{68.18} & \num{53.63} & \num{32.80} & \num{2.74} & \num{84.84} & \num{95.85} & \num{77.61}\\
      DL MobileNetV2* & \num{94.68} & \num{84.12} & \num{0.00} & \num{59.09} & \num{43.91} & \num{22.39} & \num{3.75} & \num{79.75} & \num{94.35} & \num{68.83}\\
      uNetX* & \num{89.80} & \num{77.15} & \num{0.00} & \num{42.36} & \num{24.27} & \num{13.59} & \num{0.00} & \num{75.43} & \num{91.16} & \num{45.70}\\
      Homography & \num{77.32} & \num{75.78} & \num{0.07} & \num{4.27} & \num{8.56} & \num{8.55} & \num{0.38} & \num{37.06} & \num{89.74} & \num{0.00}\\
    \end{tabu}
  \end{center}
  \label{tab:IoU}
\end{table*}

\section{Experimental Setup}

In order to evaluate the methodology presented before, we train the neural networks entirely on simulated data. In the following, we present the synthetic dataset and the training setup.

\subsection{Data Acquisition}

The data used to train and assess our proposed methodology is created in the simulation environment \textit{Virtual Test Drive~(VTD)}~\cite{Neumann-CoselEtAl_VirtualTestDrive_2009}. A recording toolchain allows the generation of potentially arbitrarily many sample images including their corresponding label.

In the simulation, the ego vehicle is equipped with four identical virtual wide-angle cameras covering a full \ang{360} surround view. Ground truth data is provided by a virtual drone camera. The BEV ground truth image is centered above the ego vehicle and has an approximate field of view of \( \SI{70}{m} \times \SI{44}{m} \). 

Both input and ground truth images are recorded at a resolution of \( \SI{964}{px} \times \SI{604}{px} \). All virtual cameras produce both realistic and semantically segmented images. For semantic segmentation, nine different semantic classes are considered for the visible areas (\textit{road}, \textit{sidewalk}, \textit{person}, \textit{car}, \textit{truck}, \textit{bus}, \textit{bike}, \textit{obstacle}, \textit{vegetation}).

As a trade-off between keeping simulation time low and maximizing data variety, images are recorded at \SI{2}{Hz}. In total, the dataset contains approximately \num{33000} samples for training and \num{3700} samples for validation, where each sample is a set of multiple input images and one ground truth label. As we only require our method to operate in specified spatial areas, the static elements in the simulated world (i.e.\ \textit{roads}, \textit{buildings}, etc.) remain the same between training and validation data.

In order to later test a real-world application of our methods, a second synthetic dataset is recorded for usage with a single front camera. In this scenario, only three classes are considered for visible areas (\textit{road}, \textit{vehicle}, \textit{occupied space}) and only the area in front of the vehicle is of interest. For this reason, the ground truth images are left-aligned with the ego vehicle. The second dataset contains approximately \num{32000} samples for training and \num{3200} samples for validation.

\subsection{Training Setup}

To keep training and inference time relatively short, network input images and target labels are center-cropped to an aspect ratio of 2:1 and resized to a resolution of \( \SI{512}{px} \times \SI{256}{px} \). The input images are converted to a one-hot representation. In order to counter class imbalance in the dataset, the loss function is modified to weigh semantic classes according to the logarithm of their relative occurrence. During training, the Adam optimizer with a learning rate of 
\num{1e-4} and parameters \( \beta_1 = 0.9 \) and \( \beta_2 = 0.999 \) is applied to batches of size \num{5}.

\subsection{Evaluation Metrics}

The \textit{Intersection-over-Union}~(IoU) score is used as the main metric for model performance on the task of predicting a certain semantic class. Class IoU scores are averaged into a single \textit{Mean Intersection-over-Union}~(MIoU) score.

\section{Results and Discussion}\label{sect:ResultsAndDiscussion}

In this section, we compare the performance of our method variations to each other and discuss the overall improvements of our methodology compared to the classical IPM technique. The standard homography image obtained by IPM is used as the baseline for our evaluation.

We present results for the two single-input models \textit{DeepLab Xception} and \textit{DeepLab MobileNetV2} as well as the multi-input model \textit{uNetXST}. In order to quantify the benefit of incorporating homographies into our approach, we also present results for alternative model versions without IPM. For the model of our first method variation, \textit{DeepLab}, this means to simply concatenate the multiple input images along their channel dimension, as explained in the beginning of~\sect{sec:Methodology}. For the \textit{uNetXST} model, this means to ablate the \textit{Spatial Transformer} units. In the following, these simplified models are denoted by an asterisk~(*).

Additionally, we qualitatively test the hypothesis that the proposed methodology can generalize from simulated to real-world data.

\subsection{Results on Synthetic Data}

\setcounter{table}{0}
\begin{table}[t]
  \caption{MIoU Scores (\%) on the Validation Set}
  \begin{center}
    \begin{tabular}{l c}
      \textbf{Model} & \textbf{MIoU}\\
      \hline
      \\[-1em]
      uNetXST & \textbf{\num{71.92}}\\
      DeepLab Xception & \num{71.35}\\
      DeepLab MobileNetV2 & \num{66.60}\\
      DeepLab Xception* & \num{60.13}\\
      DeepLab MobileNetV2* & \num{55.09}\\
      uNetX* & \num{45.95}\\
      Homography & \num{30.17}\\
    \end{tabular}
  \end{center}
  \label{tab:MIoU}
\end{table}

The performance of our models compared to the baseline is reported in~\tab{tab:MIoU}.

The \textit{uNetXST} model achieves the highest MIoU score on the validation set. This is the case even though \textit{uNetXST} contains substantially fewer trainable parameters than \textit{DeepLab Xception}, which is the second best performing network. The result can be seen as evidence for the hypothesis that the approach using \textit{uNetXST} benefits from being able to extract features from the non-transformed camera images, before perspective errors are introduced by IPM.

The results of the ablation study of omitting IPM from our approach (*) suggest that the erroneous homography view can indeed help to improve performance. Compared to the homography baseline itself, our proposed approach generally achieves a considerably higher performance. The values indicate that both variations of our method can successfully improve the results obtained by IPM for environment perception.

In order to further analyze the performance on a class basis, we present the respective class IoU scores in~\tab{tab:IoU}.

\begin{figure*}[!t]
  \captionsetup[subfigure]{skip=0pt}
  \begin{subfigure}[b]{0.2425\textwidth}
    \caption*{Front camera}
    \includegraphics[width=\textwidth, height=0.5\textwidth]{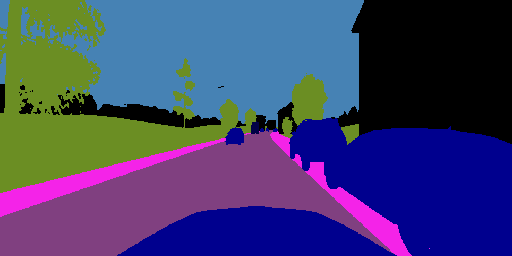}
  \end{subfigure}
  \hfill
  \begin{subfigure}[b]{0.2425\textwidth}
    \caption*{Rear camera}
    \includegraphics[width=\textwidth, height=0.5\textwidth]{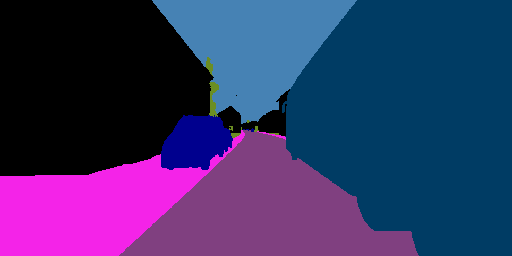}
  \end{subfigure}
  \hfill
  \begin{subfigure}[b]{0.2425\textwidth}
    \caption*{Left camera}
    \includegraphics[width=\textwidth, height=0.5\textwidth]{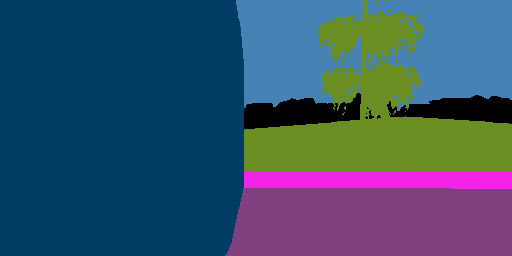}
  \end{subfigure}
  \hfill
  \begin{subfigure}[b]{0.2425\textwidth}
    \caption*{Right camera}
    \includegraphics[width=\textwidth, height=0.5\textwidth]{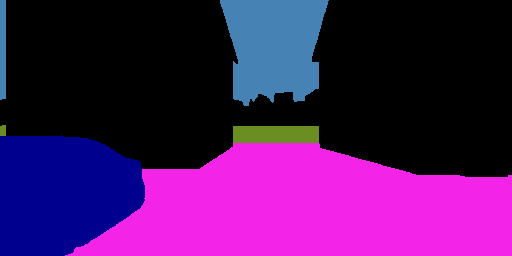}
  \end{subfigure}%

  \vspace{0.5\baselineskip}
  \captionsetup[subfigure]{skip=2pt}
  \begin{subfigure}[b]{0.2425\textwidth}
    \includegraphics[width=\textwidth, height=0.5\textwidth]{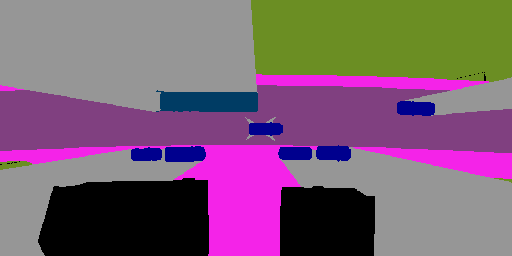}
    \caption*{Ground Truth}
  \end{subfigure}
  \hfill
  \begin{subfigure}[b]{0.2425\textwidth}
    \includegraphics[width=\textwidth, height=0.5\textwidth]{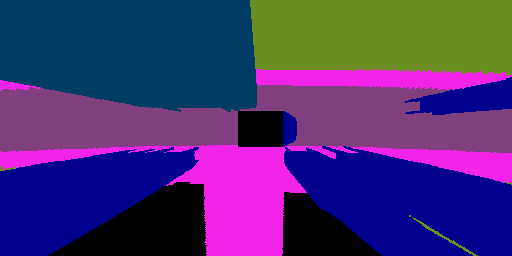}
    \caption*{Homography}
  \end{subfigure}
  \hfill
  \begin{subfigure}[b]{0.2425\textwidth}
    \includegraphics[width=\textwidth, height=0.5\textwidth]{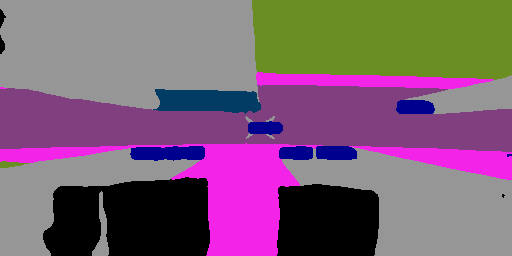}
    \caption*{DeepLab Xception}
  \end{subfigure}
  \hfill
  \begin{subfigure}[b]{0.2425\textwidth}
    \includegraphics[width=\textwidth, height=0.5\textwidth]{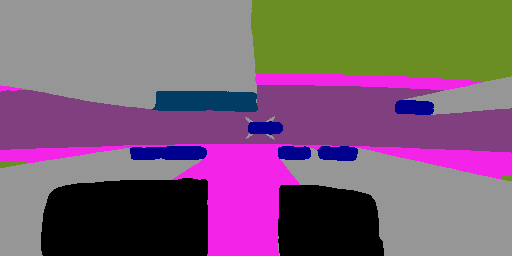}
    \caption*{uNetXST}
  \end{subfigure}%

  \vspace{0.5\baselineskip}
  \captionsetup[subfigure]{skip=0pt}
  \begin{subfigure}[b]{0.2425\textwidth}
    \caption*{Front camera}
    \includegraphics[width=\textwidth, height=0.5\textwidth]{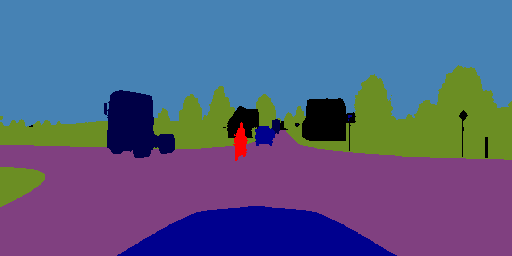}
  \end{subfigure}
  \hfill
  \begin{subfigure}[b]{0.2425\textwidth}
    \caption*{Rear camera}
    \includegraphics[width=\textwidth, height=0.5\textwidth]{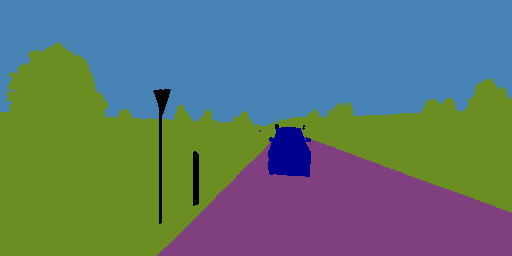}
  \end{subfigure}
  \hfill
  \begin{subfigure}[b]{0.2425\textwidth}
    \caption*{Left camera}
    \includegraphics[width=\textwidth, height=0.5\textwidth]{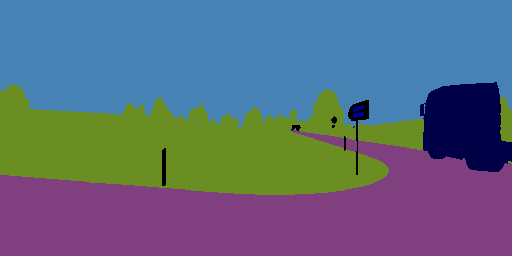}
  \end{subfigure}
  \hfill
  \begin{subfigure}[b]{0.2425\textwidth}
    \caption*{Right camera}
    \includegraphics[width=\textwidth, height=0.5\textwidth]{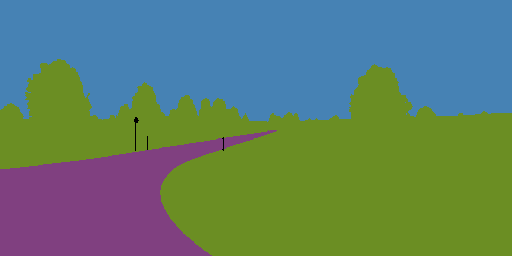}
  \end{subfigure}%

  \vspace{0.5\baselineskip}
  \captionsetup[subfigure]{skip=2pt}
  \begin{subfigure}[b]{0.2425\textwidth}
    \includegraphics[width=\textwidth, height=0.5\textwidth]{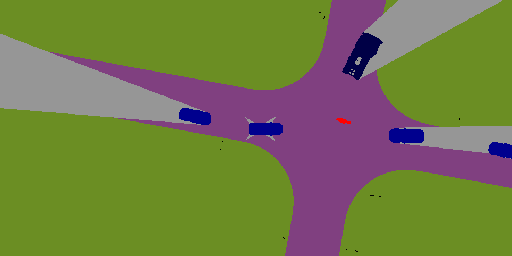}
    \caption*{Ground Truth}
  \end{subfigure}
  \hfill
  \begin{subfigure}[b]{0.2425\textwidth}
    \includegraphics[width=\textwidth, height=0.5\textwidth]{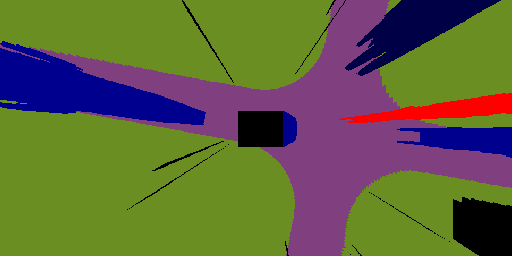}
    \caption*{Homography}
  \end{subfigure}
  \hfill
  \begin{subfigure}[b]{0.2425\textwidth}
    \includegraphics[width=\textwidth, height=0.5\textwidth]{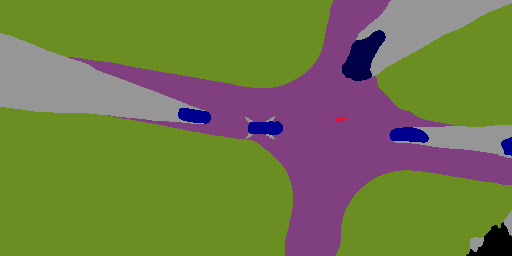}
    \caption*{DeepLab Xception}
  \end{subfigure}
  \hfill
  \begin{subfigure}[b]{0.2425\textwidth}
    \includegraphics[width=\textwidth, height=0.5\textwidth]{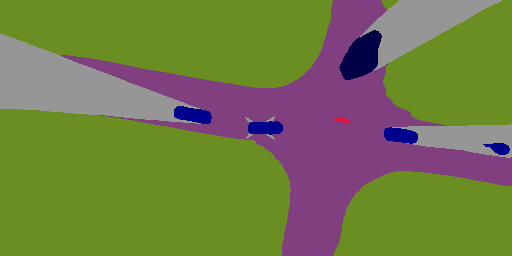}
    \caption*{uNetXST}
  \end{subfigure}%
  \caption{Example results on simulated data from the validation set}
  \label{fig:Results}
\end{figure*}

All three proposed networks perform best on the prediction of semantic classes covering large areas, e.g.\ \textit{road} and \textit{vegetation}. Good IoU scores are achieved for \textit{cars}, \textit{trucks}, and \textit{buses}, which are all dynamic traffic participants. All models struggle with the correct prediction and localization of \textit{bikes} and especially \textit{persons}. This can be attributed to the fact that both classes represent small objects in BEV and also show the least occurrence in the training dataset. The \textit{uNetXST} results for \textit{bikes} and \textit{persons} indicate that the method can indeed profit from processing the raw and non-transformed camera images. Further measures to counter class imbalance, apart from a weighted loss function, could improve the results on these two classes. The models without IPM~(*) consistently perform worse than their counterparts.

A qualitative comparison between our two method variations and the baseline can be made by analyzing the examples depicted in~\fig{fig:Results}. For both exemplary scenes, we present the input images of the four vehicle-mounted cameras, the ground truth image, the homography image, and predictions from our \textit{DeepLab Xception} and \textit{uNetXST} approaches.

The errors introduced by IPM's flat world assumption are clearly visible in the homography images. Our two models perform well at computing the correct BEV of the scene.

For the first example, moving and parked vehicles are localized particularly well and the predicted object dimensions closely match the ground truth data. The occlusion shadows are reasonably cast and are intercepted by the detection of the two buildings. Note that in contrast to \textit{uNetXST}, the \textit{DeepLab Xception} model cannot reliably infer the building dimensions from the homography image. The second example poses another challenging scene at a 4-way intersection with cars, a truck and a motorcycle. Our results show a good localization of traffic participants. The estimation of object dimensions seems worse compared to the first example. However, due to the intersection, the vehicles are slightly rotated to each other, which is not the case for most of the training samples. Note that the rightmost car is almost completely occluded and thus not properly detected.

Compared to the homography image, both variations of our approach successfully eliminate errors introduced by IPM. Additionally, they reasonably predict areas in BEV, which are occluded from the vehicle camera perspective.

\subsection{Real-World Application}

\begin{figure*}[!t]
  \begin{subfigure}[b]{0.2425\textwidth}
    \includegraphics[width=\textwidth, height=0.5\textwidth]{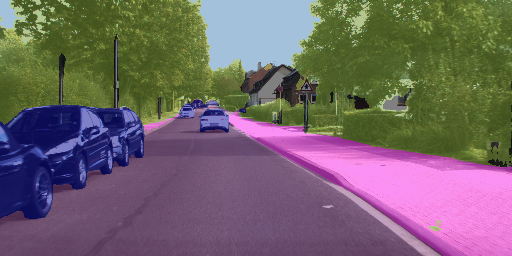}
  \end{subfigure}
  \hfill
  \begin{subfigure}[b]{0.2425\textwidth}
    \includegraphics[width=\textwidth, height=0.5\textwidth]{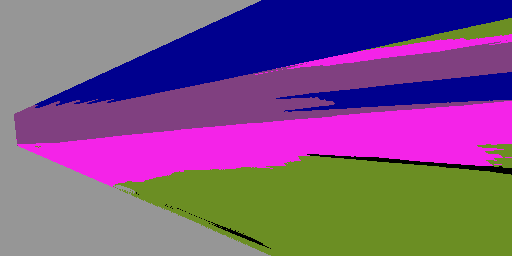}
  \end{subfigure}
  \hfill
  \begin{subfigure}[b]{0.2425\textwidth}
    \includegraphics[width=\textwidth, height=0.5\textwidth]{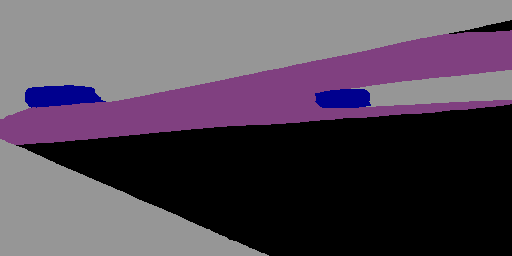}
  \end{subfigure}
  \hfill
  \begin{subfigure}[b]{0.2425\textwidth}
    \includegraphics[width=\textwidth, height=0.5\textwidth]{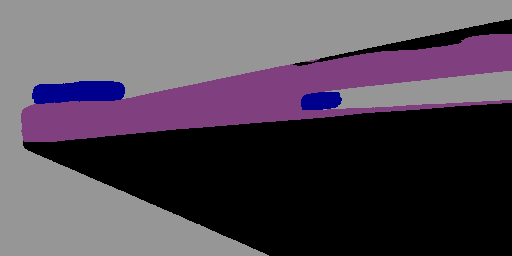}
  \end{subfigure}%

  \vspace{0.5\baselineskip}
  \captionsetup[subfigure]{skip=2pt}
  \begin{subfigure}[b]{0.2425\textwidth}
    \includegraphics[width=\textwidth, height=0.5\textwidth]{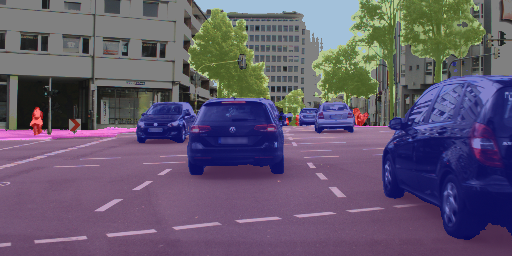}
    \caption*{Front camera with sem. segmentation}
  \end{subfigure}
  \hfill
  \begin{subfigure}[b]{0.2425\textwidth}
    \includegraphics[width=\textwidth, height=0.5\textwidth]{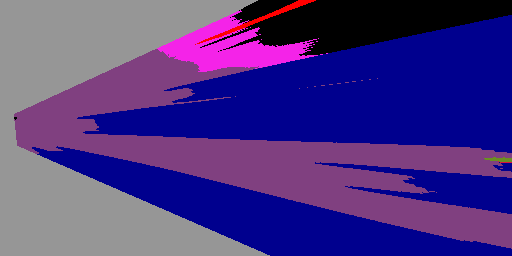}
    \caption*{Homography}
  \end{subfigure}
  \hfill
  \begin{subfigure}[b]{0.2425\textwidth}
    \includegraphics[width=\textwidth, height=0.5\textwidth]{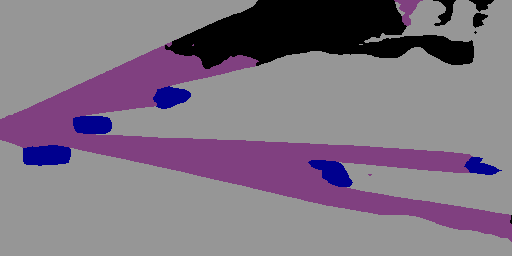}
    \caption*{DeepLab Xception}
  \end{subfigure}
  \hfill
  \begin{subfigure}[b]{0.2425\textwidth}
    \includegraphics[width=\textwidth, height=0.5\textwidth]{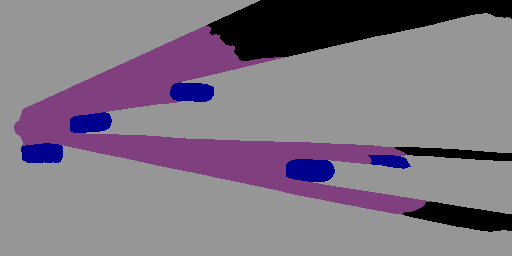}
    \caption*{uNetXST}
  \end{subfigure}%
  \caption{Example results from real-world application}
  \label{fig:RealWorldApplication}
\end{figure*}

In order to test our methodology on real-world data, we need a way of obtaining semantically segmented camera images as input to our approach. To this end, we employ an extra CNN for semantic segmentation that achieves a MIoU score of \num{79.56}{\%} on an internally labeled testing dataset.

The BEVs of two real-world scenes as computed by our \textit{DeepLab Xception} and \textit{uNetXST} models are shown in~\fig{fig:RealWorldApplication}. Both make reasonable predictions for the location and dimension of other traffic participants, but the \textit{uNetXST} model produces smoother and qualitatively better results.

In the first example, both networks reasonably predict the positions and dimensions of the parked vehicles on the left and the car ahead. In the second example, all five visible vehicles, even partially occluded ones, are detected by both models. Here, \textit{uNetXST} generally produces more reasonable object dimensions, especially for the more distant vehicles on the right.

Note that due to vehicle dynamics, in reality, a vehicle camera's pose relative to the road plane is not constant, as was the case for the simulated data. The fixed IPM transformation used for both models could thus be miscalibrated in the scenes depicted in~\fig{fig:RealWorldApplication}. Measuring vehicle dynamics and incorporating dynamic transformation changes into the network inference could therefore still improve the results in the real world.

\section{Conclusion}

We have proposed a methodology capable of transforming the images of multiple vehicle-mounted cameras to semantically segmented images in bird's eye view. In the process, errors resulting from the incorrect flatness assumption underlying \textit{Inverse Perspective Mapping} are removed. The usage of synthetic datasets and an input abstraction to semantically segmented representations of the camera images allows the application to real-world data without manual labeling of BEV images. Additionally, our method is able to accurately predict occluded areas in BEV images. We have designed the neural network \textit{uNetXST}, which processes multiple inputs and employs in-network transformations. This way the network is able to outperform popular architectures such as \textit{DeepLab Xception} on the task. All models trained using our approach quantitatively and qualitatively outperform the results obtained by only applying \textit{Inverse Perspective Mapping}.

Further research is motivated by the potential contribution the presented methodology can make to environment perception via cameras. One promising idea is to incorporate further input such as depth information. Depth information could be computed from stereo cameras, estimated by approaches for monocular camera depth estimation, or obtained from sensors such as LiDAR. Regarding a real-world application, the approach needs to be tested with a 360\(^{\circ}\) multi-camera setup, which will require good semantic segmentation performance not only on front camera images.


\bibliographystyle{IEEEtran}
\bibliography{references}

\begin{thebibliography}{10}
\providecommand{\url}[1]{#1}
\csname url@samestyle\endcsname
\providecommand{\newblock}{\relax}
\providecommand{\bibinfo}[2]{#2}
\providecommand{\BIBentrySTDinterwordspacing}{\spaceskip=0pt\relax}
\providecommand{\BIBentryALTinterwordstretchfactor}{4}
\providecommand{\BIBentryALTinterwordspacing}{\spaceskip=\fontdimen2\font plus
\BIBentryALTinterwordstretchfactor\fontdimen3\font minus
  \fontdimen4\font\relax}
\providecommand{\BIBforeignlanguage}[2]{{%
\expandafter\ifx\csname l@#1\endcsname\relax
\typeout{** WARNING: IEEEtran.bst: No hyphenation pattern has been}%
\typeout{** loaded for the language `#1'. Using the pattern for}%
\typeout{** the default language instead.}%
\else
\language=\csname l@#1\endcsname
\fi
#2}}
\providecommand{\BIBdecl}{\relax}
\BIBdecl

\bibitem{MallotEtAl_InversePerspectiveMapping_1991}
H.~A. Mallot, H.~H. B{\"u}lthoff, J.~J. Little, and S.~Bohrer, ``Inverse
  perspective mapping simplifies optical flow computation and obstacle
  detection,'' \emph{Biological Cybernetics}, vol.~64, pp. 177--185, 1991.

\bibitem{BarHillelEtAl_RecentProgressRoad_2014}
A.~Bar~Hillel, R.~Lerner, D.~Levi, and G.~Raz, ``Recent progress in road and
  lane detection: a survey,'' \emph{Machine Vision and Applications}, vol.~25,
  no.~3, pp. 727--745, 2014.

\bibitem{ZhaoEtAl_MultiSourceDomain_2019}
S.~Zhao, B.~Li, X.~Yue, Y.~Gu, P.~Xu, R.~Hu, H.~Chai, and K.~Keutzer,
  ``{Multi-source Domain Adaptation for Semantic Segmentation},'' in
  \emph{Advances in Neural Information Processing Systems}, 2019, pp.
  7285--7298.

\bibitem{SungEtAl_DevelopmentImageSynthesis_2012}
K.~Sung, J.~Lee, J.~An, and E.~Chang, ``Development of {{Image Synthesis
  Algorithm}} with {{Multi}}-{{Camera}},'' in \emph{2012 {{IEEE}} 75th
  {{Vehicular Technology Conference}} ({{VTC Spring}})}.\hskip 1em plus 0.5em
  minus 0.4em\relax {IEEE}, 2012, pp. 1--5.

\bibitem{ZhangEtAl_SurroundViewCamera_2014}
B.~Zhang, V.~Appia, I.~Pekkucuksen, Y.~Liu, A.~U. Batur, P.~Shastry, S.~Liu,
  S.~Sivasankaran, and K.~Chitnis, ``A {{Surround View Camera Solution}} for
  {{Embedded Systems}},'' in \emph{2014 {{IEEE Conference}} on {{Computer
  Vision}} and {{Pattern Recognition Workshops}}}.\hskip 1em plus 0.5em minus
  0.4em\relax {IEEE}, 2014, pp. 676--681.

\bibitem{BrulsEtAl_RightAngledPerspective_2019}
T.~Bruls, H.~Porav, L.~Kunze, and P.~Newman, ``The {{Right}} ({{Angled}})
  {{Perspective}}: {{Improving}} the {{Understanding}} of {{Road Scenes Using
  Boosted Inverse Perspective Mapping}},'' in \emph{2019 {{IEEE Intelligent
  Vehicles Symposium}} ({{IV}})}, 2019, pp. 302--309.

\bibitem{ZhuEtAl_GenerativeAdversarialFrontal_2019}
X.~Zhu, Z.~Yin, J.~Shi, H.~Li, and D.~Lin, ``Generative {{Adversarial Frontal
  View}} to {{Bird View Synthesis}},'' \emph{arXiv:1808.00327 [cs]}, 2019.

\bibitem{Schmidhuber_MakingWorldDifferentiable_1990}
{J{\"u}rgen Schmidhuber}, ``{Making the World Differentiable: On Using
  Self-Supervised Fully Recurrent Neural Networks for Dynamic Reinforcement
  Learning and Planning in Non-Stationary Environments},'' Tech. Rep., 1990.

\bibitem{GoodfellowEtAl_GenerativeAdversarialNets_2014}
I.~J. Goodfellow, J.~{Pouget-Abadie}, M.~Mirza, B.~Xu, D.~{Warde-Farley},
  S.~Ozair, A.~Courville, and Y.~Bengio, ``Generative {{Adversarial Nets}},''
  in \emph{Proceedings of the 27th {{International Conference}} on {{Neural
  Information Processing Systems}} -- NIPS'14}, vol.~2.\hskip 1em plus 0.5em
  minus 0.4em\relax {MIT Press}, 2014, pp. 2672--2680.

\bibitem{PalazziEtAl_LearningMapVehicles_2017}
A.~Palazzi, G.~Borghi, D.~Abati, S.~Calderara, and R.~Cucchiara, ``Learning to
  {{Map Vehicles}} into {{Bird}}'s {{Eye View}},'' in \emph{Image {{Analysis}}
  and {{Processing}} - {{ICIAP}} 2017}, vol. 10484.\hskip 1em plus 0.5em minus
  0.4em\relax {Cham}: {Springer International Publishing}, 2017, pp. 233--243.

\bibitem{RoddickEtAl_OrthographicFeatureTransform_2019}
T.~Roddick, A.~Kendall, and R.~Cipolla, ``Orthographic {{Feature Transform}}
  for {{Monocular 3D Object Detection}},'' in \emph{British {{Machine Vision
  Conference}} ({{BMVC}})}, 2019.

\bibitem{SenguptaEtAl_AutomaticDenseVisual_2012}
S.~Sengupta, P.~Sturgess, L.~Ladicky, and P.~H.~S. Torr, ``Automatic dense
  visual semantic mapping from street-level imagery,'' in \emph{2012
  {{IEEE}}/{{RSJ International Conference}} on {{Intelligent Robots}} and
  {{Systems}}}, 2012, pp. 857--862.

\bibitem{Dziubinski_SemanticSegmentationSemantic_2019}
\BIBentryALTinterwordspacing
M.~Dziubi{\'n}ski. (2019, 05) From semantic segmentation to semantic bird's-eye
  view in the {{CARLA}} simulator. [Online]. Available:
  \url{https://medium.com/asap-report/from-semantic-segmentation-to-semantic-birds-eye-view-in-the-carla-simulator-1e636741af3f}
\BIBentrySTDinterwordspacing

\bibitem{KaehlerBradski_LearningOpenCVComputer_2017}
A.~Kaehler and G.~R. Bradski, \emph{Learning {{OpenCV}} 3: {{Computer Vision}}
  in {{C}}++ with the {{OpenCV}} Library}, 1st~ed.\hskip 1em plus 0.5em minus
  0.4em\relax {O'Reilly Media}, 2017.

\bibitem{ChenEtAl_EncoderDecoderAtrousSeparable_2018}
L.-C. Chen, Y.~Zhu, G.~Papandreou, F.~Schroff, and H.~Adam,
  ``Encoder-{{Decoder}} with {{Atrous Separable Convolution}} for {{Semantic
  Image Segmentation}},'' in \emph{Computer {{Vision}} \textendash{} {{ECCV}}
  2018}, vol. 11211.\hskip 1em plus 0.5em minus 0.4em\relax {Springer
  International Publishing}, 2018, pp. 833--851.

\bibitem{SandlerEtAl_MobileNetV2InvertedResiduals_2018}
M.~Sandler, A.~Howard, M.~Zhu, A.~Zhmoginov, and L.-C. Chen, ``{{MobileNetV2}}:
  {{Inverted Residuals}} and {{Linear Bottlenecks}},'' in \emph{2018
  {{IEEE}}/{{CVF Conference}} on {{Computer Vision}} and {{Pattern
  Recognition}}}.\hskip 1em plus 0.5em minus 0.4em\relax {IEEE}, 2018, pp.
  4510--4520.

\bibitem{Chollet_XceptionDeepLearning_2017}
F.~Chollet, ``Xception: {{Deep Learning}} with {{Depthwise Separable
  Convolutions}},'' in \emph{2017 {{IEEE Conference}} on {{Computer Vision}}
  and {{Pattern Recognition}}}.\hskip 1em plus 0.5em minus 0.4em\relax {IEEE},
  2017, pp. 1800--1807.

\bibitem{RonnebergerEtAl_UNetConvolutionalNetworks_2015}
O.~Ronneberger, P.~Fischer, and T.~Brox, ``U-{{Net}}: {{Convolutional
  Networks}} for {{Biomedical Image Segmentation}},'' in \emph{Medical {{Image
  Computing}} and {{Computer}}-{{Assisted Intervention}} \textendash{}
  {{MICCAI}} 2015}, vol. 9351.\hskip 1em plus 0.5em minus 0.4em\relax {Springer
  International Publishing}, 2015, pp. 234--241.

\bibitem{JaderbergEtAl_SpatialTransformerNetworks_2015}
M.~Jaderberg, K.~Simonyan, A.~Zisserman, and K.~Kavukcuoglu, ``Spatial
  {{Transformer Networks}},'' in \emph{Proceedings of the 28th {{International
  Conference}} on {{Neural Information Processing Systems}}}, vol.~2.\hskip 1em
  plus 0.5em minus 0.4em\relax {MIT Press}, 2015, pp. 2017--2025.

\bibitem{Neumann-CoselEtAl_VirtualTestDrive_2009}
K.~von {Neumann-Cosel}, M.~Dupuis, and C.~Weiss, ``Virtual {{Test Drive}}
  \textendash{} {{Provision}} of a {{Consistent Tool}}-{{Set}} for
  [{{D}},{{H}},{{S}},{{V}}]-in-the-{{Loop}},'' in \emph{Proceedings of
  {{Driving Simulation Conference Europe}}, 2009}, 2009.

\end{thebibliography}

\end{document}